\documentclass{article}

\usepackage{arxiv}

\usepackage[utf8]{inputenc} 
\usepackage[T1]{fontenc}    
\usepackage{hyperref}       
\usepackage{url}            
\usepackage{booktabs}       
\usepackage{amsfonts}       
\usepackage{nicefrac}       
\usepackage{microtype}      
\usepackage{lipsum}
\usepackage{tikz}
\usepackage{pgfplots}
\usepackage{amsmath}
\usepackage{subcaption}
\usepackage{adjustbox}
\usepackage{multicol}
\usepackage{multirow}

\pgfplotsset{compat = newest}

\newcommand{\argmin}{\operatornamewithlimits{argmin}}
\newcommand{\argmax}{\operatornamewithlimits{argmax}}

\title{Semantics-Preserving Adversarial Training}

\author{
  Wonseok Lee, Hanbit Lee, Sang-goo Lee \\ 
  Department of Computer Science and Engineering\\
  Seoul National University\\
  Seoul, Korea \\
  \texttt{\{wslee, skcheon, sglee\}@europa.snu.ac.kr} \\
}

\begin{document}
\maketitle

\begin{abstract}
Adversarial training is a defense technique that improves adversarial robustness of a deep neural network (DNN) by including adversarial examples in the training data.
In this paper, we identify an overlooked problem of adversarial training in that these adversarial examples often have different semantics than the original data, introducing unintended biases into the model.
We hypothesize that such non-semantics-preserving (and resultingly ambiguous) adversarial data harm the robustness of the target models.
To mitigate such unintended semantic changes of adversarial examples, we propose \textit{semantics-preserving adversarial training} (SPAT) which encourages perturbation on the pixels that are shared among all classes when generating adversarial examples in the training stage.
Experiment results show that SPAT improves adversarial robustness and achieves state-of-the-art results in CIFAR-10 and CIFAR-100.
\end{abstract}


\section{Introduction}
\label{sec:intro}

Recent successes in many deep learning applications such as computer vision \cite{he2016deep}, speech recognition \cite{graves2013speech}, game playing \cite{silver2016mastering}, and natural language processing \cite{devlin2018bert} raised expectations for AI applications in real life.
However, as Deep Neural Networks (DNNs) turn out to be too brittle and susceptible to small perturbations known as adversarial examples \cite{szegedy2013intriguing, goodfellow2014explaining}, serious concerns are being raised on applying DNNs to safety-critical real life tasks such as face recognition \cite{parkhi2015deep}, autonomous driving \cite{chen2015deepdriving}, and medical applications \cite{finlayson2019adversarial}.

A broad definition of an adversarial example is an input to a machine learning model that is intentionally designed by an attacker to fool the model into producing an incorrect output \cite{goodfellow_papernot_definition}.
In the image classification domain, although unrestricted attacks such as adversarial rotations, and translations \cite{engstrom2019exploring} exist, typically, adversarial examples are crafted by adding some small perturbations to examples to change model outputs, where perturbation size is restricted by an $L_p$ norm  $\epsilon$-ball constraint.
These are called sensitivity-based adversarial examples \cite{tramer2020fundamental}.
Underlying assumption here is that every data point inside an $\epsilon$-ball is semantically identical.
Extensive studies were made to effectively find adversarial examples inside an $\epsilon$-ball and to make classifiers empirically or provably robust to such $L_p$ norm bounded adversarial attacks \cite{ross2018improving, madry2018towards, zhang2019theoretically, wang2019improving, wong2018provable, cohen2019certified, balunovic2019adversarial}.
However, many defense methods including even the recent researches were later shown to be ineffective \cite{carlini2017adversarial, athalye2018obfuscated, uesato2018adversarial, tramer2020adaptive}.

In spite of such bitter failures, adversarial training, which incorporates adversarial examples into the training data, remains as one of the best defense methods.
Projected gradient descent (PGD) is typically utilized to find adversarial examples used in training stage.
PGD finds a data point $x'$ which is most likely to be adversarial inside an $\epsilon$-ball centered at the original data $x$ by maximizing the loss function and the $x'$ is used as training data in place of the original data $x$.
Therefore, adversarial training can be thought of as an online data augmentation technique.

Recently, authors in \cite{tramer2020fundamental} exposed a problem of adversarial training.
This failure mode motivates us to rethink the adversarial training from the beginning.
The problem of adversarial training is that actually, data points in $\epsilon$-ball are not always semantically identical.
There are perturbations that change oracle (human) label inside $\epsilon$-ball in MNIST dataset \cite{tramer2020fundamental}.
Additionally, adversarial examples of adversarially trained models are often perceived as samples from different classes \cite{tsipras2018robustness}.
Even if the label does not change, at least the semantics can become mixed or ambiguous.
In the perspective of the data augmentation, such a data is undesirable because it makes data noisy and disrupts model from learning intended semantics.
Instead of learning the intended task-relevant information, a model learns unintended features and wrong type of invariances.
We hypothesize that such non-semantics-preserving (and resultingly ambiguous) adversarial data harm the robustness of the target model.

To mitigate such unintended semantic changes of adversarial examples, we propose \textit{semantics-preserving adversarial training} (SPAT) which encourages perturbation on the pixels that are shared among all classes when generating adversarial examples in the training stage.
We show in Section \ref{sec:attack} that perturbing on the pixels that are shared among all classes is more effective in preserving original semantics than perturbing on the pixels that are only influential to the true class.
Our aim is to train a model with more semantics-preserving adversarial examples.

By proposing SPAT, we are arguing for the necessity to separate adversarial examples for training and adversarial examples for evaluating the robustness.
When evaluating the adversarial robustness, even if the semantics is mixed or ambiguous, it is plausible to decide the label of the data based on the dominant semantics of the data as long as the label of the data is same.
However, when training, such semantically ambiguous data disturbs a model from learning intended semantics.

SPAT is a simple yet effective method.
It is worth noting that SPAT is orthogonal to existing adversarial training variants in that SPAT suggests a new method for generating adversarial examples used in training stage which remains relatively unexplored.
We show in Section \ref{sec:experiments} that when combined with TRADES and MART, SPAT achieves state-of-the-art results in CIFAR-10 and CIFAR-100 and is further improved with additional unlabeled data.

Our contributions are summarized as:
\begin{itemize}
    \item We analyze and visualize adversarial examples on various settings in a complex dataset.
    \item We identify an overlooked problem of adversarial training in that these adversarial examples often have different semantics than the original data, introducing unintended biases into the model.
    To mitigate such unintended semantic changes of adversarial examples, we propose \textit{semantics-preserving adversarial training} (SPAT).
    \item We experimentally show that SPAT can improve adversarial robustness and achieve state-of-the-art results in CIFAR-10 and CIFAR-100.
\end{itemize}

\section{Preliminaries}
\label{sec:prelim}
Consider a standard classification task.
Given a classfier which is parametrized by $\theta$ and data $(x, y) \sim D$ where $x$ is a image and $y \in \{0, 1\}^K$ is a one-hot encoded class label:

\paragraph{Standard Training}
The goal of Empirical Risk Minimization (ERM) is to find parameter $\theta$ that minimizes the risk:

\begin{equation}
\label{eq:erm}
    \theta^* = \argmin_\theta E_{(x,y) \sim D}[ L(\theta, x, y) ]
\end{equation}

\paragraph{Adversarial Training}
In adversarial training, we allow some perturbations for each data point $x$.
The perturbation set $S$ is chosen to capture semantic similarity of images.
Usually, $l_p$ ball around $x$ is used:

\begin{equation}
\label{eq:ball}
    B_\epsilon^p(x) = \{ x': ||x - x'||_p \le \epsilon \}
\end{equation}

In this paper, we use $p=\infty$.
Then, adversarial risk minimization is formulated as a saddle point problem:
\begin{equation}
\label{eq:adv_risk_mini}
    \theta^* = \argmin_\theta E_{(x,y) \sim D}[ \max_{x' \in B_\epsilon^\infty (x)} L(\theta, x', y) ]
\end{equation}

which can be rewritten as:
\begin{equation}
\label{eq:argmin}
    \theta^* = \argmin_\theta E_{(x,y) \sim D}[ L(\theta, \hat{x}', y) ]
\end{equation}
where
\begin{equation}
\label{eq:argmax}
    \hat{x}' =  \argmax_{x' \in B_\epsilon^\infty (x)} L(\theta, x', y)
\end{equation}
It is alternating between inner maximization problem and outer minimization problem and rewritten formulation can be thought of as just a modification of ERM where adversarial data is used instead of natural data.
In its variants, loss function $L$ in equation \ref{eq:argmin} and equation \ref{eq:argmax} are not essentially same.

\paragraph{Projected Gradient Descent}
To approximately solve inner maximization problem, standard adversarial training uses Projected Gradient Descent (PGD) \cite{madry2018towards}.
First, we introduce Fast Gradient Sign Method (FGSM) \cite{goodfellow2014explaining} and then move on to PGD.
FGSM finds an adversarial example as

\begin{equation}
    x' = x + \epsilon sgn(\nabla_x L(\theta, x, y) )
\end{equation}

which is a one-step method. The multi-step variant of FGSM is called PGD:
\begin{equation}
    x^{(t+1)} = \Pi_{B_\epsilon^\infty(x)}(x^{(t)} + \alpha sgn(\nabla_{x^{(t)}} L(\theta, x^{(t)}, y) ))
\end{equation}
where $\alpha > 0$ is a step size and $\Pi$ is a projection operator that projects adversarial example back to $\epsilon$-ball centered at original data point $x$ and $x^{(t)}$ is a adversarial example at step $t$.
For both methods, at the beginning($x^{(0)}$), small Gaussian or uniform noise may be added to $x$ and that is called random start.
PGD is performed for fixed iteration $T$ and is called PGD-T algorithm.
Most commonly used function for surrogate loss of inner maximization is standard cross entropy loss \cite{madry2018towards, wang2019improving, mao2019metric}.
KL-divergence function and other methods have been used as well \cite{zhang2019theoretically, miyato2018virtual, carlini2017towards}.
Note that $\hat{x}'$ found by PGD is not always adversarial.
Instead, PGD finds a point that is most likely to be misclassified by the model.

\section{Semantics-Preserving Adversarial Training}
\label{sec:defense}
In this section, we analyze current problem of PGD-based adversarial training and propose \textit{semantics-preserving adversarial training} (SPAT) algorithm, 
which encourages perturbation on the pixels that are shared among all classes when generating adversarial examples in the training stage.

\subsection{Problem of PGD-training}

\begin{figure}
  \centering
  \includegraphics[width=\columnwidth]{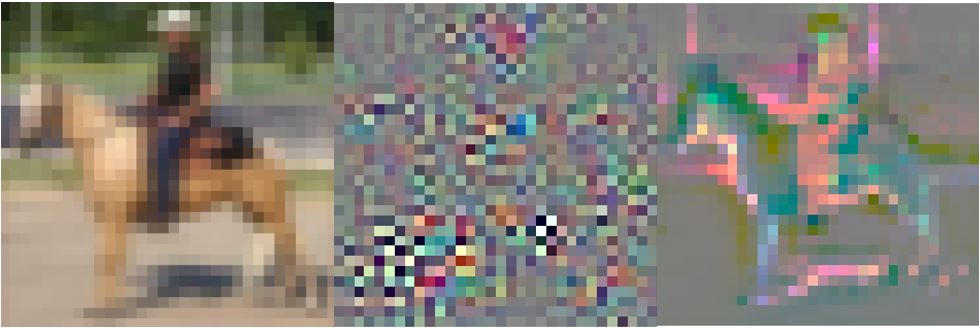}
  \caption{(Left) natural image (Middle) gradient w.r.t. x of standard model (Right) gradient w.r.t. x of adversarially trained model.}
  \label{fig:fig3_1}
\end{figure}

Several researches have been conducted on how the survived adversarially trained models differ from the standard models \cite{zhang2019interpreting, tsipras2018robustness}.
Specifically, authors in \cite{tsipras2018robustness} has shown that the gradients of adversarially trained models align well with perceptually relevant features of the input image while the gradients of standard models seem as mere noises to humans.
See Figure \ref{fig:fig3_1} for examples.

Since PGD finds adversarial examples based on the gradients of the models, the distinct aspects of gradients induce significant disparity between the PGD-generated examples from the adversarially trained models and the standard models.
Adversarial examples of standard models seem as noisy version of the original images.
In contrast, adversarial examples of adversarially trained models look semantically different from the original images and they often belong to different classes.
See Figure \ref{fig:4_2_epsilon} for examples.

Moreover, there are perturbations that change oracle (human) label inside $\epsilon$-ball in MNIST dataset \cite{tramer2020fundamental}.
This may apply to other datasets and other size of $\epsilon$-balls as well.
Thus it can not be guaranteed that adversarial examples have same semantics as original images.
If not totally change labels, adversarial perturbations may make images ambiguous by adding semantics of different classes or by erasing semantics of the original classes.

In the perspective of an adversarial attack whose goal is to generate images that the model misclassifies, this is a very interesting phenomenon and it is a evidence that adversarial training can teach the semantics to the models to some degree.
However, in the perspective of a data augmentation in adversarial training,
training with such an attack is harmful because it 
prevents the model from learning intended semantics in that it mixes up the task-relevant and task-irrelevant information.

We hypothesize that such non-semantics-preserving (and resultingly ambiguous) adversarial data harm the robustness of the target model and this may be one of the cause of phenomenon that defenses against sensitivity-based attacks harm a model's accuracy on invariance-based attacks \cite{tramer2020fundamental}.
That is, making the model robust in $\epsilon$-balls actually gives the model invariance in wrong direction so that the model becomes invariant to semantics.

\subsection{Semantics-Preserving Adversarial Training}
To solve this problem, we propose \textit{semantics-preserving adversarial training} (SPAT), where we use label smoothed cross entropy loss (LSCE) \cite{szegedy2016rethinking} instead of standard cross entropy loss (CE) for surrogate loss of inner maximization problem.
That is, we use 

\begin{equation}
\label{eq:lsce}
LSCE(p, y) = \sum_{k=1}^K -y_k^{LS} \log(p_k) \\
\end{equation}
where
\begin{equation}
\label{eq:ykls}
y_k^{LS} = 
\begin{cases}
(1-\alpha) & \text{if} \; y_k = 1 \\
\alpha/(K-1) & \text{if} \; y_k = 0
\end{cases}
\end{equation}

for surrogate loss of inner maximization where $\alpha \in [0, 1]$ is a label smoothing hyperparameter and $p_k$ is k-th element of softmax layer output.
For full formulation, refer to Equation \ref{eq:lspgd}.
Note that LSCE is equivalent to CE when $\alpha = 0$.
Since PGD with cross entropy loss perturbs towards increasing loss only with original class, it encourages erasing semantics of true class and adding semantics of other classes.
As a result, PGD with cross entropy loss changes the original semantics of the images.

\begin{equation}
\label{eq:lspgd}
    \hat{x}' =  \argmax_{x' \in B_\epsilon^\infty (x)} LSCE(p(x', \theta), y)
\end{equation}

In contrast, as SPAT encourages to perturb on the pixels that are shared among all classes, it mitigates two causes of semantic changes of PGD-generated adversarial examples: adding semantics of other classes and erasing semantics of the original class.
Such a semantics-preserving effect increases as label smoothing hyperparameter $\alpha$ gets bigger.
As $\alpha$ gets bigger, PGD will perturb more on parts that are common across all other classes, therefore lesser erasing semantics of the true class.
However, as $\alpha$ gets bigger, it provides less invariance to the model since evenly distributed loss prevents the sample from diverging from the original data point.
Therefore, there is a tradeoff.
When using LSCE loss for PGD, we call it PGD-LS for convenience and same go for PGD-CE and PGD-KL.

\begin{equation}
\label{eq:kl_xent}
    CE(p, q) = Entropy(p) + D_{KL}(p || q)
\end{equation}

Since CE loss function is equivalent to KL divergence except for the entropy (which is the constant part), if the softmax probability is same, LSCE loss is equivalent to KL divergence (refer to Equation \ref{eq:kl_xent}).
However, we claim that LSCE has advantage over KL divergence in that we are able to control how much semantics to preserve with label smoothing hyperparameter $\alpha$.
Since KL-div highly depends on the sample prediction computed by trained models, it varies from model to model and from example to example.
In contrast, with LSCE, we are able to control the ratio between the true class and other classes.
Overall, PGD-LS can be thought of as a generalization of PGD-CE and PGD-KL.

\subsection{Combining with Adversarial Training Variants}
Since our method is changing the surrogate loss for inner maximization problem, it is orthogonal to various existing adversarial training methods.
Therefore, we combine our method with Madry \cite{madry2018towards}, TRADES \cite{zhang2019theoretically}, and MART \cite{wang2019improving}.

\paragraph{Madry + SPAT}
Loss function is formulated as $CE(p(\hat{x}', \theta), y)$.
\paragraph{TRADES + SPAT}
Loss function is formulated as $CE(p(x, \theta), y) + KL(p(x, \theta) || p(\hat{x}', \theta))$.
\paragraph{MART + SPAT}
Loss function is formulated as $BCE(p(\hat{x}', \theta), y) + KL(p(x, \theta) || p(\hat{x}', \theta))(1-p_y(x, \theta))$.

All the adversarial examples $\hat{x}'$ is generated by Equation \ref{eq:lspgd}.

\section{Analysis of Adversarial Examples}
\label{sec:attack}
In this section, we compare our proposed PGD-LS attack with various PGD-based attacks.
First, we show that semantic changes occur in $\epsilon$-balls and such semantic changes can be mitigated with PGD-LS.
Next, to show the effect of hyperparameter $\alpha$ and compare with other PGD-based attacks numerically, we plot attack success rate curves on a standard model and an adversarially trained model.

\subsection{Visualizing Various Adversarial Examples}
\begin{figure}[h]
  \centering
  \includegraphics[scale=0.3]{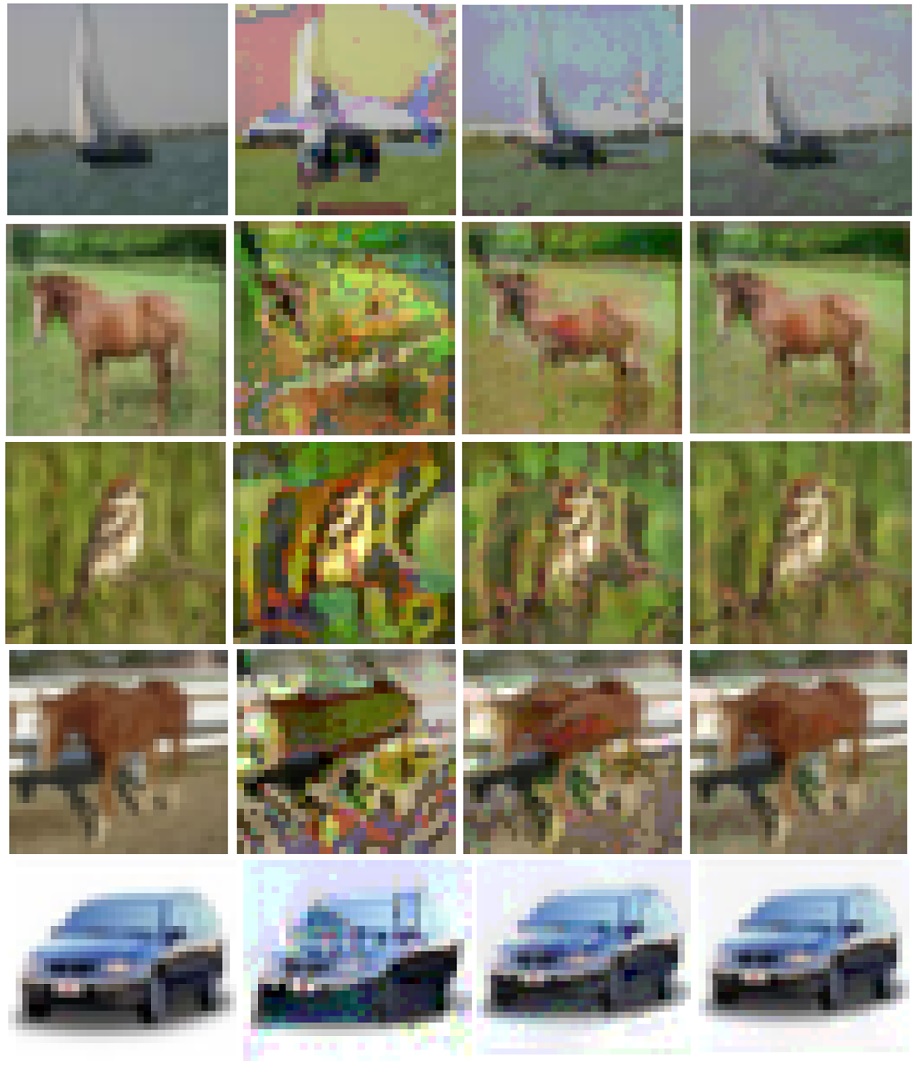}
  \caption{Adversarial examples of adversarially trained model generated on various perturbation limits. All adversarial images are generated by C\&W$_\infty$ attack. From left to right: original image, $\epsilon=32/255$, $\epsilon=16/255$, $\epsilon=8/255$. From top to bottom: ship to airplane, horse to frog, bird to frog, horse to frog, automobile to ship.}
  \label{fig:4_2_epsilon}
\end{figure}

Here, we visualize adversarial examples generated by various PGD-based attacks and various perturbation limits.
First, to test the effect of perturbation limit on adversarial examples, we generate adversarial examples with C\&W$_\infty$ attack on various perturbation limits on CIFAR-10 dataset.
Figure \ref{fig:4_2_epsilon} shows the result.
With $\epsilon=32/255$, labels of the images completely change.
For example, images in first row show a ship turning into a airplane.
On $\epsilon=16/255$, semantics of the images change to some degree and labels often become ambiguous and mixed.
For instance, images in first row show a ship becoming ambiguous between a ship and a airplane and images in second row show that the shape of a horse is deformed.
On $\epsilon=8/255$, which is the most commonly used perturbation limit on CIFAR-10 dataset, labels of the images are preserved but some images show mixed semantics.
For example, semantics of adversarial image in third row is mixed but the label is preserved.
However, since we cannot inspect every image in every used dataset, we cannot assure that there is no label-changing or ambiguous adversarial examples in defined epsilon balls.

\begin{figure}[h]
  \centering
  \includegraphics[scale=0.3]{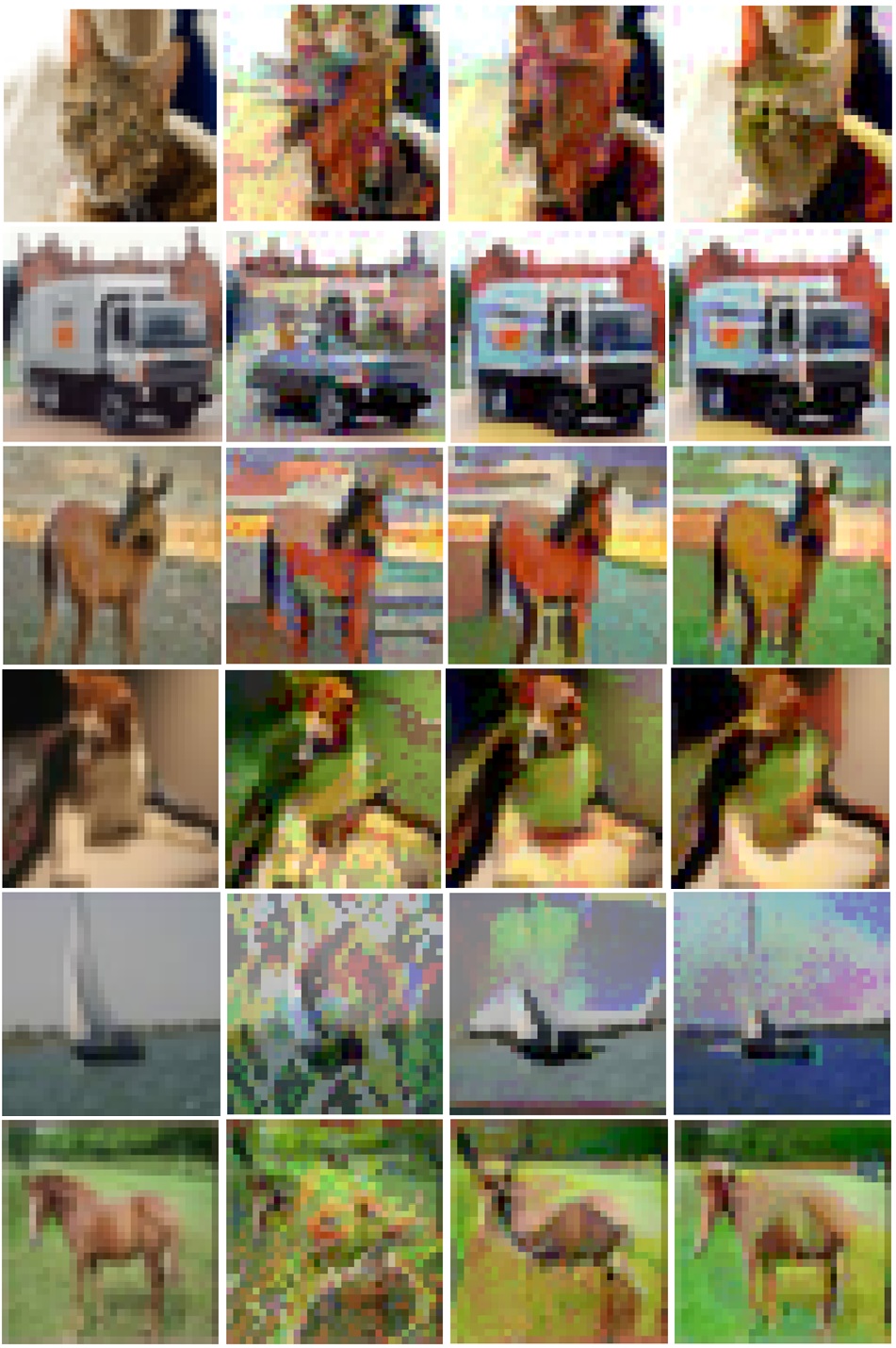}
  \caption{Adversarial examples generated on adversarially trained model with various attacks. All adversarial images are generated on $\epsilon=32/255$. From left to right: original image, PGD-CE, PGD-LS ($\alpha=0.2$), PGD-LS ($\alpha=0.8$).}
  \label{fig:4_2_attack}
\end{figure}

Secondly, to confirm that PGD-LS attack is more effective at preserving semantics than PGD-CE attack, we visualize adversarial examples generated by PGD-CE and PGD-LS attack on perturbation limit of $\epsilon=32/255$.
Figure \ref{fig:4_2_attack} shows that adversarial examples generated by PGD-LS attack preserve more semantics than adversarial examples generated by PGD-CE attack and semantics-preserving effect is greater with larger $\alpha$.
Therefore, by using larger $\alpha$ on larger perturbation limit, adversarial training can become more stable by a larger semantics-preserving effect.

\subsection{Comparing the Attack Success Rate}
\begin{figure}[ht]
    \centering

    \includegraphics[width=\columnwidth]{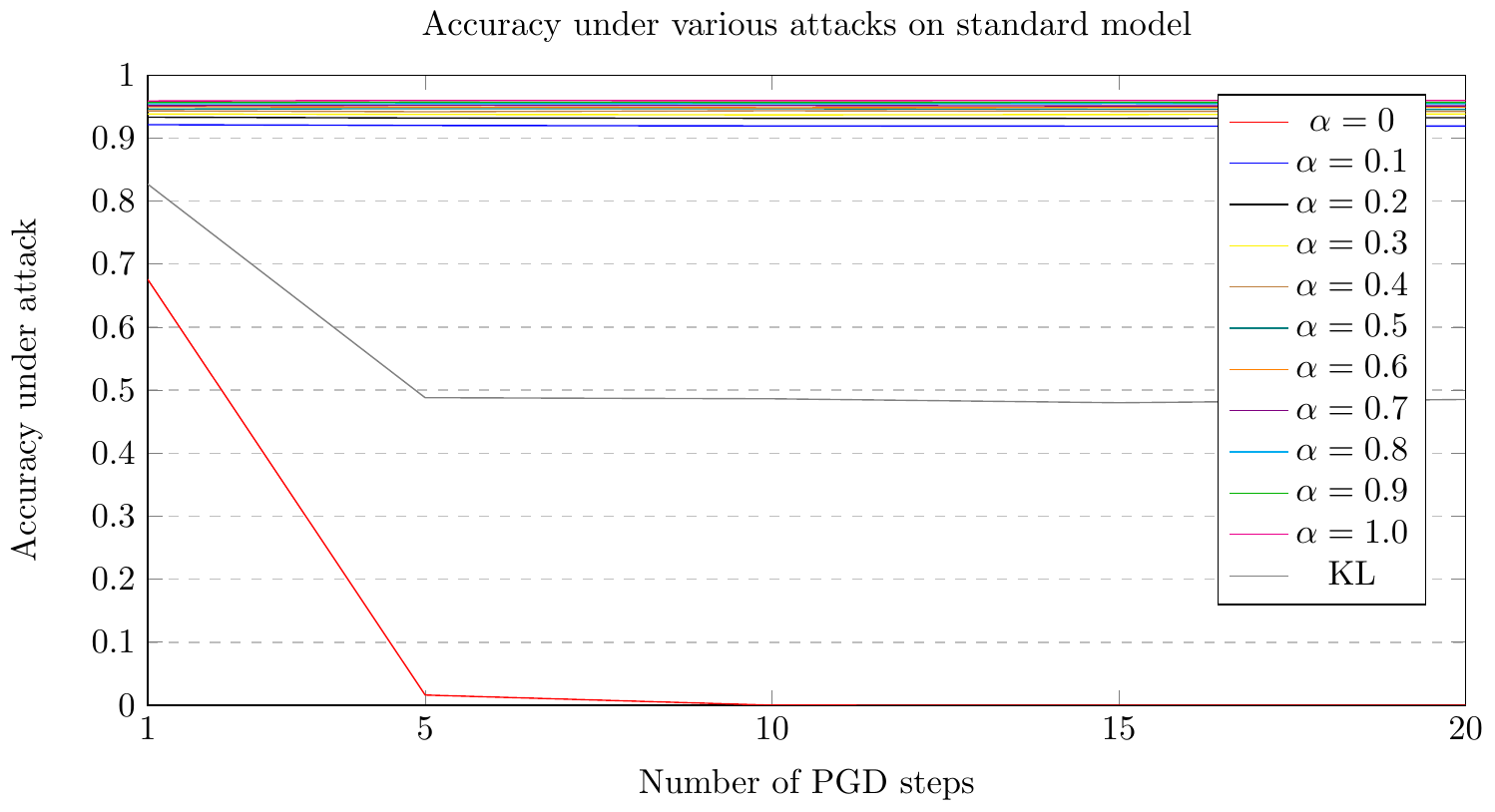}
    \includegraphics[width=\columnwidth]{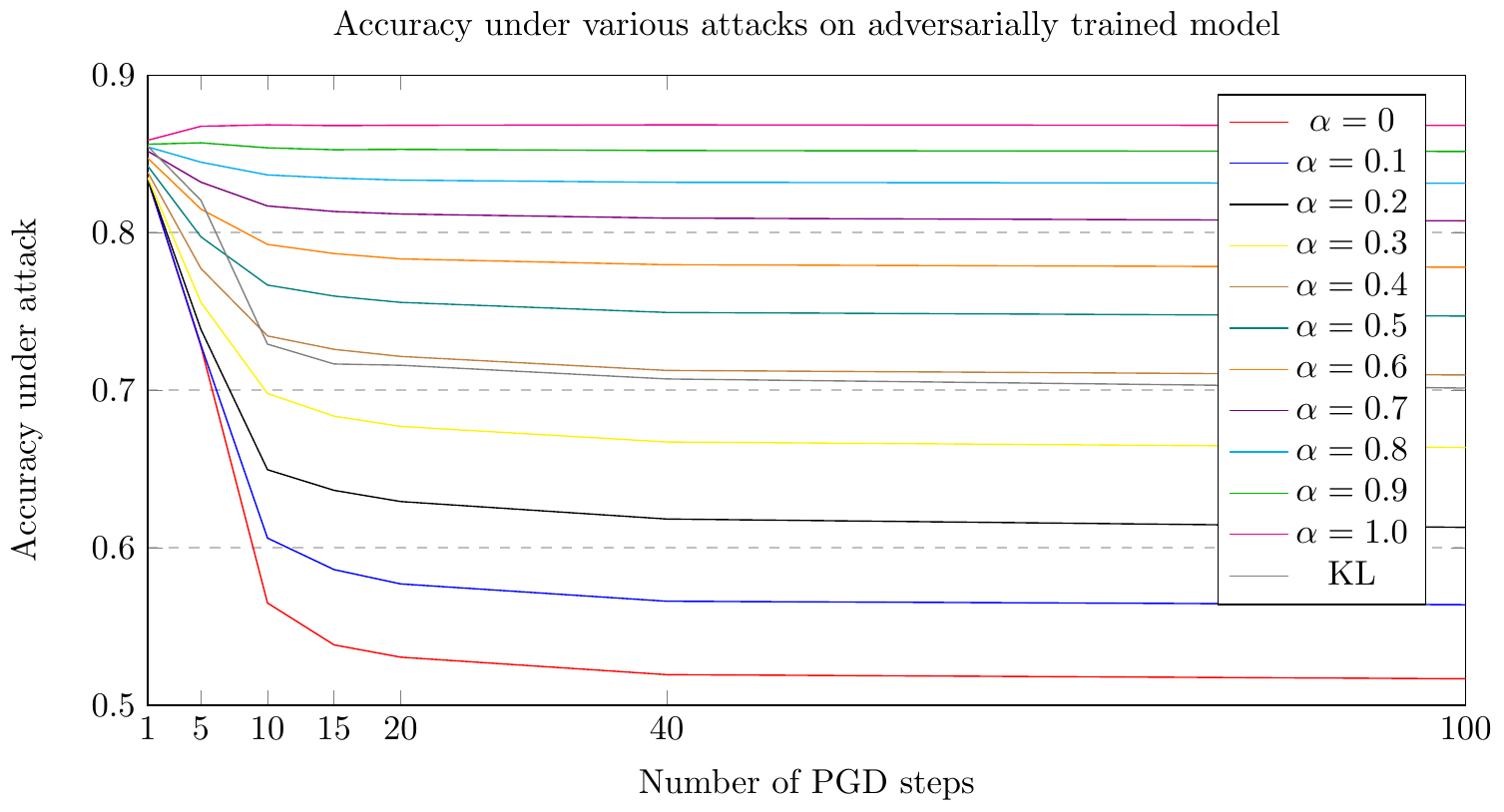}

\caption{Robust accuracy under various attacks on a standard model and an adversarially trained model.}
\label{fig:4_1}
\end{figure}

Here, we analyze the effect of label smoothing hyperparameter $\alpha$ on attack success rate of PGD-LS against a standard model and an adversarially trained model (Madry).
We vary $\alpha$ from $0$ to $1$ with stride $0.1$.
Note that when $\alpha=0$, it is equivalent to PGD-CE.
We also plot PGD-KL \cite{zhang2019theoretically} for comparison.
Note that robust accuracy is equal to $1-$ attack success rate.
Figure \ref{fig:4_1} shows the result.

As expected, we observe that bigger $\alpha$ leads to lower attack success rate (= higher robust accuracy) due to its larger semantics-preserving effect.
It is worth noting that for both model, when $\alpha$ is $1.0$, accuracy under attacks get higher than accuracy for clean examples.
This is because of closed set nature of classification problem.
In closed set classification, moving away from every class except true class results in moving towards true class.

Attack success rates in a standard model and an adversarially trained model show quite different aspect.
On a standard model, $\alpha=0$ shows huge difference from other PGD-LS.
In contrast, on adversarially trained model, PGD-LS shows gradual changes.

\paragraph{Relation of PGD-KL vs PGD-LS vs PGD-CE}
We also notice that PGD-KL shows difference in a standard model and an adversarially trained model.
Therefore, we investigate what $\alpha$ value leads PGD-LS to have similar attack success rate with PGD-KL.
We show results in Table \ref{tab:kl_nat} and \ref{tab:kl_adv}.
Results show that for a standard model, PGD-LS with $\alpha$ between 1e-4 and 5e-5 is similar with PGD-KL and for an adversarially trained model, PGD-LS with $\alpha$ between $0.4$ and $0.5$ is similar with PGD-KL.
Probably standard training makes a highly confident classifier whereas adversarial training yields less confident classifier.
The disadvantage of PGD-KL is that since its power of attack (or attack success rate) is determined by the model's sample prediction which is uncontrollable, it is undependable.

\begin{table}[ht]
  \centering
  \begin{tabular}{l|l|lllll}
    \toprule
    \multirow{2}{4em}{Attack}& \multirow{2}{4em}{PGD-KL} & \multicolumn{5}{c}{PGD-LS} \\
         &  & 1e-2 & 1e-3 & 1e-4 & 5e-5 & 1e-5  \\
    \midrule
    PGD-20 acc & 47.51  & 80.04    &   69.23   & 51.28 & 44.16 & 25.3\\
    \bottomrule
  \end{tabular}
  \caption{Robust accuracy (\%) of standard model under PGD-KL and PGD-LS attack with different $\alpha$ values.}
  \label{tab:kl_nat}
\end{table}

\begin{table}[ht]
  \centering
  \begin{tabular}{l|l|lllll}
    \toprule
    \multirow{2}{4em}{Attack}& \multirow{2}{4em}{PGD-KL} & \multicolumn{5}{c}{PGD-LS} \\
        &  & 0.1 & 0.2 & 0.3 & 0.4 & 0.5  \\
    \midrule
    PGD-20 acc & 71.58  & 57.7  & 62.93 & 67.7 & 72.14 & 75.57\\
    \bottomrule
  \end{tabular}
  \caption{Robust accuracy (\%) of adversarially trained model under PGD-KL and PGD-LS attack with different $\alpha$ values.}
  \label{tab:kl_adv}
\end{table}

\section{Experiments \& Results}
\label{sec:experiments}
In this section, we first verify the efficacy of SPAT empirically by several experiments and then check how the label smoothing parameter $\alpha$ of the SPAT affects the accuracy of a classifier on various perturbation limits.

\subsection{Evaluating Robustness}
We train WideResNet-34-10 \cite{zagoruyko2016wide} on CIFAR-10 and CIFAR-100 dataset \cite{krizhevsky2009learning}
to benchmark state-of-the-art robustness and train with 500k unlabeled data on CIFAR-10 to achieve further improvements.

\subsubsection{CIFAR-10 \& CIFAR-100}
We compare our method with adversarial training variants: 1) Madry \cite{madry2018towards}, 2) TRADES \cite{zhang2019theoretically}, and 3) MART \cite{wang2019improving}.

\paragraph{Training Details}
For CIFAR-10, we follow all the settings in MART.
Models are trained with SGD with momentum $0.9$, weight decay 7e-4 and initial learning rate is $0.1$ and divided by $0.1$ at 75-th and 90-th epoch.
All images are normalized into [0, 1] and when training, data augmentation such as random horizontal flipping and random crop with 4 pixel padding is performed.
The perturbation limit is $\epsilon=8/255$ and for training attack, we use PGD-10 with random start and step size is $\epsilon/4$.
For all hyperparameters, we use $\lambda=6$ for TRADES and TRADES + SPAT, $\lambda=5$ for MART and MART + SPAT.
For CIFAR-100, we use same settings except for weight decay which follow their original implementations.

We test all models against FGSM(w/o random start), PGD-20, and C\&W$_\infty$ (optimized by PGD for 30 steps) \cite{carlini2017towards} attacks.

\begin{table}[ht]
  \centering
  \begin{tabular}{l|llll|llll}
    \toprule
    & \multicolumn{4}{c|}{CIFAR-10} & \multicolumn{4}{c}{CIFAR-100} \\
    Defense Methods & Natural & FGSM & PGD-20 & CW$_\infty$ & Natural & FGSM & PGD-20 & CW$_\infty$  \\
    \midrule
    Madry   & 84.39             & 59.93             & 56.37 & 54.14 & \textbf{61.77}    & 34.42 & 31.51 & 30.10 \\
    TRADES  & \textbf{85.97}    & \textbf{62.29}    & 57.32 & 54.38 & 57.13             & 32.90 & 31.02 & 28.07 \\
    MART    & 83.70              & 61.93             & 58.46 & 53.28 & 58.56             & 36.46 & 34.12 & 30.16 \\
    \midrule
    TRADES + SPAT ($\alpha=0.1$) & 84.60 & 61.46 & 58.22             & \textbf{54.97} & 55.54 & 33.35 & 31.20  & 28.14 \\
    MART + SPAT ($\alpha=0.3, 0.2$)   & 81.93 & 61.87 & \textbf{59.59}    & 51.57 & 60.28 & \textbf{36.91} & \textbf{34.66}  & \textbf{30.93} \\
    \bottomrule
  \end{tabular}
    \caption{Natural and Robust accuracy (\%) of WRN-34-10 trained on CIFAR-10 and CIFAR-100 dataset. MART + SPAT uses $\alpha=0.3$ on CIFAR-10 and $\alpha=0.2$ on CIFAR-100.}
    \label{tab:cifar10}
\end{table}

\paragraph{Results \& Discussion}
Table \ref{tab:cifar10} shows the result.
In CIFAR-10, our proposed method MART + SPAT outperforms all other methods in terms of PGD-20 accuracy, which is the most common comparison setting.
Also, TRADES + SPAT outperforms all other methods in terms of CW$_\infty$ accuracy.
Compared with its original algorithm TRADES, TRADES + SPAT improves on PGD-20 and CW$_\infty$ accuracy.
MART + SPAT improves on PGD-20 over MART but worsens on CW$_\infty$ accuracy.
This phenomenon is similar to relation of MART and TRADES.
MART has higher PGD-20 accuracy than TRADES but shows lower CW$_\infty$ accuracy.
We presume that this is because BCE loss used in MART (instead of CE in TRADES) sometimes cause mismatch between robustness against PGD attack and C\&W attack.
All our method worsens in natural accuracy, which conforms with claim that robustness may be inherently at odds with natural accuracy \cite{tsipras2018robustness, zhang2019theoretically}.

In CIFAR-100, our proposed method MART + SPAT outperforms all other methods in all robust accuracy and also improves natural accuracy over its original algorithm, MART.
TRADES + SPAT improves on all robust accuracy over TRADES but worsens natural accuracy.

Overall, experiment results show that our proposed method SPAT consistently improves robust accuracy.
It was generally thought that increasing the power of attack by increasing the number of attack iterations can create more robust model \cite{madry2018towards, xie2020smooth, xie2019intriguing}.
Our results show that stronger attack is not the only way to creating more robust models.
This conforms with our intuition that semantics-preserving data augmentation is important.

\subsubsection{CIFAR-10 with 500K Unlabeled Data}
Here, we investigate the additional benefit of unlabeled data with SPAT.
We follow exact same settings in RST \cite{carmon2019unlabeled}.
Specifically, we train RST + SPAT and MART + SPAT on WideResNet-28-10 and compare them with RST and MART on natural accuracy and PGD-20 (settings in \cite{carmon2019unlabeled}) accuracy.
Evaluation results are shown in Table \ref{tab:cifar_unlabel}.
Results show that SPAT improves PGD-20 accuracy on both RST and MART.
We again confirm that SPAT consistently improves robust accuracy.

\begin{table}
  \centering
  \begin{tabular}{lll}
    \toprule
    Defense     & Natural     & PGD-20 \\
    \midrule
    RST & 89.65  & 63.00  \\
    MART  & \textbf{89.81} & 63.06 \\
    \midrule
    RST + SPAT ($\alpha=0.1$) & 89.52 & \textbf{63.47}  \\
    MART + SPAT ($\alpha=0.1$) & 89.44 & 63.37 \\
    \bottomrule
  \end{tabular}
  \caption{Natural and Robust accuracy (\%) of WRN-28-10 trained on CIFAR-10 with 500k unlabeled dataset.}
  \label{tab:cifar_unlabel}
\end{table}

\subsection{Effect of Label Smoothing Hyperparameter $\alpha$}
To test the effect of label smoothing hyperparameter $\alpha$ on SPAT on various perturbation limits, we train ResNet-50 \cite{he2016deep} on CIFAR-10 dataset.
We apply SPAT on the standard adversarial training method, Madry \cite{madry2018towards}.
We vary $\alpha$ from 0 to 1 with stride $0.2$.
Note that when $\alpha=0$, it is equivalent to Madry.

\paragraph{Adversarial Setting}
We train models on various perturbation limits to see the effect of SPAT on various amount of possible semantics change.
The perturbation limits for training are $\epsilon=4/255, 8/255, 16/255$ and $\epsilon=8/255$ for evaluation.
For training, we use PGD-10 with random start and step size is $\epsilon/4$.
For evaluation, we use PGD-20 with random start and step size is $\epsilon/10$.

\begin{figure}[t]
\centering
\begin{subfigure}{0.49\textwidth}
    \includegraphics[]{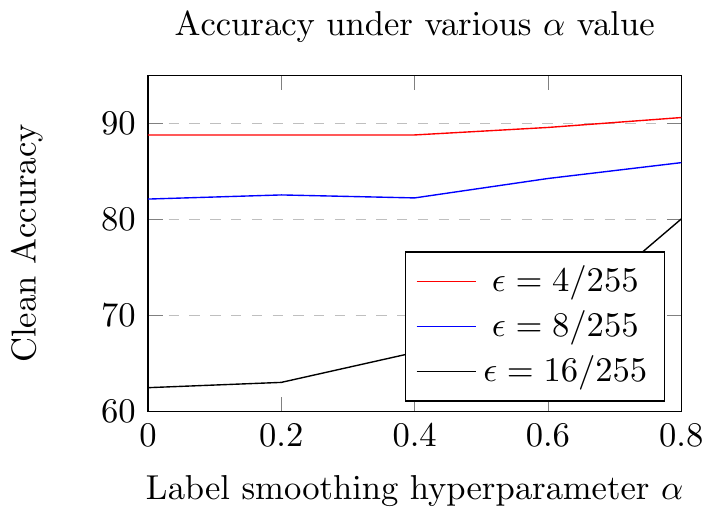}
\end{subfigure}
\begin{subfigure}{0.49\textwidth}
    \includegraphics[]{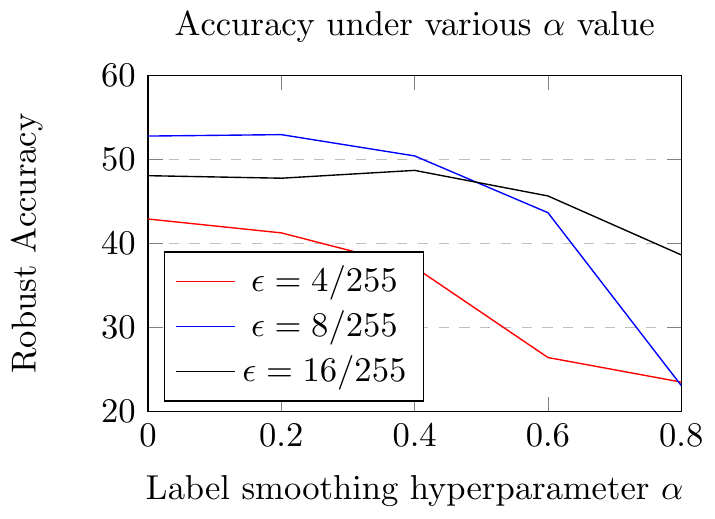}
\end{subfigure}
\caption{Clean and Robust Accuracy (\%) under various $\alpha$ and perturbation limits on CIFAR-10 dataset.}
    \label{fig:5_1}
\end{figure}

\paragraph{Evaluation Results}
In Figure \ref{fig:5_1}, we show the performance of Madry and SPAT w.r.t. $\alpha$ on various perturbation limits.
Clean accuracy refers to the accuracy of a classifier evaluated on natural images and robust accuracy refers to the accuracy of a classifier evaluated on adversarial examples generated by PGD-20.
All DNNs trained by SPAT show higher clean accuracy compared to the models that are trained by Madry.
The models trained by SPAT with higher $\alpha$ have higher clean accuracy, but when $\alpha=1$, training gets broken.
The model trained by SPAT with $\alpha=1$ has clean accuracy of $32.44\%$ and robust accuracy of $0\%$ when $\epsilon=8/255$.
We presume that training with data that is more 'friendly' than clean data (or flattering) kills training.

For robust accuracy, a DNN trained with $\alpha=0$ get highest robust accuracy when $\epsilon=4/255$.
When $\epsilon=8/255$, SPAT with $\alpha=0.2$ get highest robust accuracy.
When $\epsilon=16/255$, SPAT with $\alpha=0.4$ get highest robust accuracy.
In contrast to clean accuracy, robust accuracy peaks at certain $\alpha$ value and decreases as it gets farther away from the peak $\alpha$ value.
In addition, the peak $\alpha$ value is higher on bigger $\epsilon$-ball.

Higher clean accuracy is achieved with smaller $\epsilon$-ball.
In contrast, higher robust accuracy is achieved with middle sized $\epsilon$-ball and some semantics preservation.
Overall, this conforms with our intuition that although some degree of invariances are essential to achieve robustness, too much invariance (or unintended bias) caused by non-semantics-preserving data hinders adversarial training and that can be mitigated with semantics-preserving adversarial training.

We confirm that semantics-preserving adversarial training with proper choice of $\alpha$ helps to increase robustness of the model even with large perturbation limits.
The optimal amount of semantics to be preserved which is controlled by $\alpha$ is dependent on the radius of the $\epsilon$-ball since larger perturbation limit allows for more semantic changes.
When $\alpha$ is too big compared to $\epsilon$, SPAT makes adversarial data too close to original data so that it does not provide enough invariances and the model becomes less robust.
SPAT could serve as a guide to finding the appropriate size of $\epsilon$-ball.

\section{Related Work}
Concurrent work of \cite{zhang2020attacks} studied employing least adversarial data minimizing the loss among the adversarial data that are confidently misclassified to mitigate cross-over mixtures.
They proposed early-stopped PGD to achieve adversarial robustness without compromising the natural generalization.
On the contrary, we propose semantics-preserving adversarial training to mitigate unintended semantic changes of adversarial data which we hypothesized as a cause of degradation of adversarial robustness of the model.
Specifically, we use alternative surrogate loss for inner maximization to encourage perturbation on the pixels that are shared among all classes.
Note that our method can also be combined with early-stopped PGD.

\section{Conclusion \& Future Work}
\label{sec:conclusion}
In this paper, motivated by recently discovered vulnerability of adversarially trained DNNs,
we investigate the effect of semantics of adversarial data on adversarial robustness.
We observe that not only insufficient invariance but also too much invariance (= semantics-changing adversarial data) impairs robustness.
To mitigate such semantic changes of adversarial data for adversarial training, we propose a semantics-preserving adversarial training (SPAT) algorithm.
Experiment results show that SPAT with proper choice of $\alpha$ which is dependent on the perturbation limit improves robustness.
We leave efficiently finding the optimal combination of $\epsilon$ and $\alpha$ as a future work.

\bibliographystyle{unsrt}  
\bibliography{references}  

\begin{thebibliography}{10}

\bibitem{he2016deep}
Kaiming He, Xiangyu Zhang, Shaoqing Ren, and Jian Sun.
\newblock Deep residual learning for image recognition.
\newblock In {\em Proceedings of the IEEE conference on computer vision and
  pattern recognition}, pages 770--778, 2016.

\bibitem{graves2013speech}
Alex Graves, Abdel-rahman Mohamed, and Geoffrey Hinton.
\newblock Speech recognition with deep recurrent neural networks.
\newblock In {\em 2013 IEEE international conference on acoustics, speech and
  signal processing}, pages 6645--6649. IEEE, 2013.

\bibitem{silver2016mastering}
David Silver, Aja Huang, Chris~J Maddison, Arthur Guez, Laurent Sifre, George
  Van Den~Driessche, Julian Schrittwieser, Ioannis Antonoglou, Veda
  Panneershelvam, Marc Lanctot, et~al.
\newblock Mastering the game of go with deep neural networks and tree search.
\newblock {\em nature}, 529(7587):484--489, 2016.

\bibitem{devlin2018bert}
Jacob Devlin, Ming-Wei Chang, Kenton Lee, and Kristina Toutanova.
\newblock Bert: Pre-training of deep bidirectional transformers for language
  understanding.
\newblock {\em arXiv preprint arXiv:1810.04805}, 2018.

\bibitem{szegedy2013intriguing}
Christian Szegedy, Wojciech Zaremba, Ilya Sutskever, Joan Bruna, Dumitru Erhan,
  Ian Goodfellow, and Rob Fergus.
\newblock Intriguing properties of neural networks.
\newblock {\em arXiv preprint arXiv:1312.6199}, 2013.

\bibitem{goodfellow2014explaining}
Ian~J Goodfellow, Jonathon Shlens, and Christian Szegedy.
\newblock Explaining and harnessing adversarial examples.
\newblock {\em arXiv preprint arXiv:1412.6572}, 2014.

\bibitem{parkhi2015deep}
Omkar~M Parkhi, Andrea Vedaldi, and Andrew Zisserman.
\newblock Deep face recognition.
\newblock 2015.

\bibitem{chen2015deepdriving}
Chenyi Chen, Ari Seff, Alain Kornhauser, and Jianxiong Xiao.
\newblock Deepdriving: Learning affordance for direct perception in autonomous
  driving.
\newblock In {\em Proceedings of the IEEE International Conference on Computer
  Vision}, pages 2722--2730, 2015.

\bibitem{finlayson2019adversarial}
Samuel~G Finlayson, John~D Bowers, Joichi Ito, Jonathan~L Zittrain, Andrew~L
  Beam, and Isaac~S Kohane.
\newblock Adversarial attacks on medical machine learning.
\newblock {\em Science}, 363(6433):1287--1289, 2019.

\bibitem{goodfellow_papernot_definition}
I.~Goodfellow and N.~Papernot.
\newblock Is attacking machine learning easier than defending it?, Feb 2017.

\bibitem{engstrom2019exploring}
Logan Engstrom, Brandon Tran, Dimitris Tsipras, Ludwig Schmidt, and Aleksander
  Madry.
\newblock Exploring the landscape of spatial robustness.
\newblock In {\em International Conference on Machine Learning}, pages
  1802--1811, 2019.

\bibitem{tramer2020fundamental}
Florian Tram{\`e}r, Jens Behrmann, Nicholas Carlini, Nicolas Papernot, and
  J{\"o}rn-Henrik Jacobsen.
\newblock Fundamental tradeoffs between invariance and sensitivity to
  adversarial perturbations.
\newblock {\em arXiv preprint arXiv:2002.04599}, 2020.

\bibitem{ross2018improving}
Andrew~Slavin Ross and Finale Doshi-Velez.
\newblock Improving the adversarial robustness and interpretability of deep
  neural networks by regularizing their input gradients.
\newblock In {\em Thirty-second AAAI conference on artificial intelligence},
  2018.

\bibitem{madry2018towards}
Aleksander Madry, Aleksandar Makelov, Ludwig Schmidt, Dimitris Tsipras, and
  Adrian Vladu.
\newblock Towards deep learning models resistant to adversarial attacks.
\newblock In {\em International Conference on Learning Representations}, 2018.

\bibitem{zhang2019theoretically}
Hongyang Zhang, Yaodong Yu, Jiantao Jiao, Eric Xing, Laurent El~Ghaoui, and
  Michael~I Jordan.
\newblock Theoretically principled trade-off between robustness and accuracy.
\newblock In {\em ICML}, 2019.

\bibitem{wang2019improving}
Yisen Wang, Difan Zou, Jinfeng Yi, James Bailey, Xingjun Ma, and Quanquan Gu.
\newblock Improving adversarial robustness requires revisiting misclassified
  examples.
\newblock In {\em International Conference on Learning Representations}, 2019.

\bibitem{wong2018provable}
Eric Wong and Zico Kolter.
\newblock Provable defenses against adversarial examples via the convex outer
  adversarial polytope.
\newblock In {\em International Conference on Machine Learning}, pages
  5286--5295, 2018.

\bibitem{cohen2019certified}
Jeremy Cohen, Elan Rosenfeld, and Zico Kolter.
\newblock Certified adversarial robustness via randomized smoothing.
\newblock In {\em International Conference on Machine Learning}, pages
  1310--1320, 2019.

\bibitem{balunovic2019adversarial}
Mislav Balunovic and Martin Vechev.
\newblock Adversarial training and provable defenses: Bridging the gap.
\newblock In {\em International Conference on Learning Representations}, 2019.

\bibitem{carlini2017adversarial}
Nicholas Carlini and David Wagner.
\newblock Adversarial examples are not easily detected: Bypassing ten detection
  methods.
\newblock In {\em Proceedings of the 10th ACM Workshop on Artificial
  Intelligence and Security}, pages 3--14, 2017.

\bibitem{athalye2018obfuscated}
Anish Athalye, Nicholas Carlini, and David Wagner.
\newblock Obfuscated gradients give a false sense of security: Circumventing
  defenses to adversarial examples.
\newblock In {\em International Conference on Machine Learning}, pages
  274--283, 2018.

\bibitem{uesato2018adversarial}
Jonathan Uesato, Brendan O'Donoghue, Aaron van~den Oord, and Pushmeet Kohli.
\newblock Adversarial risk and the dangers of evaluating against weak attacks.
\newblock {\em arXiv preprint arXiv:1802.05666}, 2018.

\bibitem{tramer2020adaptive}
Florian Tramer, Nicholas Carlini, Wieland Brendel, and Aleksander Madry.
\newblock On adaptive attacks to adversarial example defenses.
\newblock {\em arXiv preprint arXiv:2002.08347}, 2020.

\bibitem{tsipras2018robustness}
Dimitris Tsipras, Shibani Santurkar, Logan Engstrom, Alexander Turner, and
  Aleksander Madry.
\newblock Robustness may be at odds with accuracy.
\newblock In {\em International Conference on Learning Representations}, 2018.

\bibitem{mao2019metric}
Chengzhi Mao, Ziyuan Zhong, Junfeng Yang, Carl Vondrick, and Baishakhi Ray.
\newblock Metric learning for adversarial robustness.
\newblock In {\em Advances in Neural Information Processing Systems}, pages
  480--491, 2019.

\bibitem{miyato2018virtual}
Takeru Miyato, Shin-ichi Maeda, Masanori Koyama, and Shin Ishii.
\newblock Virtual adversarial training: a regularization method for supervised
  and semi-supervised learning.
\newblock {\em IEEE transactions on pattern analysis and machine intelligence},
  41(8):1979--1993, 2018.

\bibitem{carlini2017towards}
Nicholas Carlini and David Wagner.
\newblock Towards evaluating the robustness of neural networks.
\newblock In {\em 2017 ieee symposium on security and privacy (sp)}, pages
  39--57. IEEE, 2017.

\bibitem{zhang2019interpreting}
Tianyuan Zhang and Zhanxing Zhu.
\newblock Interpreting adversarially trained convolutional neural networks.
\newblock In {\em International Conference on Machine Learning}, pages
  7502--7511, 2019.

\bibitem{szegedy2016rethinking}
Christian Szegedy, Vincent Vanhoucke, Sergey Ioffe, Jon Shlens, and Zbigniew
  Wojna.
\newblock Rethinking the inception architecture for computer vision.
\newblock In {\em Proceedings of the IEEE conference on computer vision and
  pattern recognition}, pages 2818--2826, 2016.

\bibitem{zagoruyko2016wide}
Sergey Zagoruyko and Nikos Komodakis.
\newblock Wide residual networks.
\newblock {\em arXiv preprint arXiv:1605.07146}, 2016.

\bibitem{krizhevsky2009learning}
Alex Krizhevsky, Geoffrey Hinton, et~al.
\newblock Learning multiple layers of features from tiny images.
\newblock 2009.

\bibitem{xie2020smooth}
Cihang Xie, Mingxing Tan, Boqing Gong, Alan Yuille, and Quoc~V Le.
\newblock Smooth adversarial training.
\newblock {\em arXiv preprint arXiv:2006.14536}, 2020.

\bibitem{xie2019intriguing}
Cihang Xie and Alan Yuille.
\newblock Intriguing properties of adversarial training at scale.
\newblock In {\em International Conference on Learning Representations}, 2019.

\bibitem{carmon2019unlabeled}
Yair Carmon, Aditi Raghunathan, Ludwig Schmidt, John~C Duchi, and Percy~S
  Liang.
\newblock Unlabeled data improves adversarial robustness.
\newblock In {\em Advances in Neural Information Processing Systems}, pages
  11192--11203, 2019.

\bibitem{zhang2020attacks}
Jingfeng Zhang, Xilie Xu, Bo~Han, Gang Niu, Lizhen Cui, Masashi Sugiyama, and
  Mohan Kankanhalli.
\newblock Attacks which do not kill training make adversarial learning
  stronger.
\newblock {\em arXiv preprint arXiv:2002.11242}, 2020.

\end{thebibliography}

\end{document}